\documentclass{ws-jtca}
\usepackage{tikz,commath}
\usepackage{color}

\begin{document}

\markboth{D. Veerababu \& Prasanta K. Ghosh}{1-D Acoustic Field Prediction using Neural Networks}

%
\catchline{}{}{}{}{}
%

\title{Prediction of acoustic field in 1-D uniform duct with varying mean flow and temperature using neural networks}

\author{Veerababu Dharanalakota\textsuperscript{\textdagger} and Prasanta K. Ghosh\footnote{Corresponding author.}\textsuperscript{\hspace{1.5mm},\hspace{0.5mm}\textdaggerdbl}}

\address{Department of Electrical Engineering, Indian Institute of Science,\\
Bengaluru, Karnataka 560012, India \\
\email{\textsuperscript{\textdagger}veerababudha@iisc.ac.in \\
\email{\textsuperscript{\textdaggerdbl}prasantg@iisc.ac.in}}}

\maketitle

\begin{history}
\received{(Day Month Year)}
\revised{(Day Month Year)}
\end{history}

\begin{abstract}
Neural networks constrained by the physical laws emerged as an alternate numerical tool. In this paper, the governing equation that represents the propagation of sound inside a one-dimensional duct carrying a heterogeneous medium is derived. The problem is converted into an unconstrained optimization problem and solved using neural networks. Both the acoustic state variables: acoustic pressure and particle velocity are predicted and validated with the traditional Runge-Kutta solver. The effect of the temperature gradient on the acoustic field is studied. Utilization of machine learning techniques such as transfer learning and automatic differentiation for acoustic applications is demonstrated. 
\end{abstract}


\tikzset{
my dash/.style={thick,dash pattern=on 7pt off 5pt,
                postaction={draw, line cap=off, line width=3pt,
                            dash pattern=on 0.1pt off 34.9pt},
                }
         }
         
\newcommand{\blueline}{\raisebox{2pt}{\tikz{\draw[-,blue,solid,line width = 1.5pt](0,0) -- (12mm,0);}}}
\newcommand{\redline}{\raisebox{1.5pt}{\tikz{\draw[-,red,my dash,line width = 1.5pt](0,0) -- (12mm,0);}}}
\newcommand{\greenline}{\raisebox{2pt}{\tikz{\draw[-,black!40!green,dashdotted,line width = 1.5pt](0,0) -- (10mm,0);}}}
\newcommand{\blackline}{\raisebox{2pt}{\tikz{\draw[-,black,dotted,line width = 1.5pt](0,0) -- (10mm,0);}}}

\section{Introduction}
Traditionally, acoustics has relied heavily on theoretical models and empirical data to understand and predict sound behaviors in various environments\cite{Morse1986,Munjal2014}. However, the advent of big data and advanced machine learning algorithms has opened new avenues for research and application, fundamentally transforming how acoustic phenomena are analyzed and utilized. Data-driven methods in acoustics leverage vast amounts of data to train models that can uncover patterns and relationships that are often too complex for traditional analytical methods. For instance, machine learning algorithms can be trained on large datasets of acoustic signals to identify and classify sounds with high precision, leading to advancements in fields such as speech\cite{Abhay2021,Srinivasan2019}, architectural acoustics\cite{Xydis2023}, environmental noise monitoring\cite{Yeh2021,Green2020}, active noise control\cite{Cha2023,Zhang2021}, and structural health monitoring\cite{Rafiee2007,Bangalore2017}. Despite these advancements, the application of data-driven methods in acoustics is not without challenges. Without high-quality and diversified training data, one cannot build a reliable machine learning model\cite{Bianco2019}. 

In scenarios where the required data are not available enough, physics-informed neural networks (PINNs) can be used as a supplement. PINNs learn the underlying physics from the governing differential equations rather than the data from either the simulations or experiments, and predict the required quantities at par with the classical analytical and numerical methods\cite{Cuomo2022}. The researchers successfully solved the Schrodinger equation\cite{Raissi2019,Radu2022}, Korteweg-de Vries equation\cite{Raissi2019,Zhou2024}, Burger’s equation\cite{Raissi2019,Mathias2022}, Poisson equation\cite{Basir2022,Maddu2022}, Navier-Stokes equations\cite{Jin2021,Oldenburg2022}, etc. 

The solution of the Helmholtz equation using PINNs can also be evidenced in the literature\cite{Alkhalifah2021,Cui2022}. However, most of the proposed formulations focus on solving the governing equations with low-fidelity physics. Very limited research is available on the application of PINNs to real-world problems where it requires the incorporation of high-fidelity physics into the modeling. This paper demonstrates the neural network-based solution to predict the acoustic field inside a one-dimensional (1-D) uniform duct carrying a fluid whose properties change with respect to position. This work finds extensive applications in the aerospace industry, especially in the design of gas turbine combustors, where the rapidly changing medium properties introduce combustion instabilities into the system through the constant feedback loop between the acoustics of the combustor and the heat release rate\cite{Heckl2022,Li2017,Karthik2000}. Predicting acoustic field in such a heterogeneous medium is a challenging task even in the 1-D settings. Researchers are able to overcome the difficulties and provided solutions by analytical means\cite{Li2017,Munjal1986,Cummings1978}. However, developing a neural network-based solution (despite its success in recent times) is still a challenging problem for the following reasons.  
\begin{arabiclist}
\item Presence of mean flow make the coefficients the governing differential equation complex-valued. Hence, acoustic pressure becomes a complex-valued function, which ultimately makes the loss function complex-valued. However, the loss function must be a real-valued function to perform the optimization procedure.  
\item Developing a framework that can predict the acoustic field at different frequencies without altering the network architecture is not a trivial process. 
\end{arabiclist}
These problems are addressed in the current work, and the development of the formulation is presented step-by-step.

The article is organized as follows: Section~\ref{Sec:2} describes the derivation of the governing equation from the fundamentals of fluid dynamics and thermodynamics. In Section~\ref{Sec:3}, a primer on PINNs and conversion of solving the derived differential equation into an optimization problem are presented. The prediction of acoustic pressure and particle velocity from the formulation developed is presented in Section~\ref{Sec:4}. The effect of the temperature gradient on the acoustic field is also studied in the same section. The article is concluded in Section~\ref{Sec:5} with final remarks and future scope.  

\section{Derivation of the governing equation} \label{Sec:2}
Let us assume that a duct is carrying a fluid whose mean temperature and mean velocity change along the axial direction. Also, assume that the fluid is inviscid and obeys the perfect gas law. The governing equation that represents the propagation of sound inside the uniform duct with heterogeneous medium properties can be derived from the fundamental equation of fluid dynamics. According to Li et al.\cite{Li2017}, the continuity and momentum equations that govern fluid flow inside the one-dimensional duct can be written as
\begin{align}
    \frac{\partial\rho}{\partial t}+\frac{\partial}{\partial x}(\rho u) &= 0, \label{Eq:1}\\
    \rho\frac{\partial u}{\partial t}+\rho u \frac{\partial u}{\partial x}+\frac{\partial p}{\partial x} &= 0, \label{Eq:2}
\end{align}
where $\rho$, $u$, and $p$ are the density, velocity, and pressure of the fluid medium, respectively. If $\rho'$, $u'$, and $p'$ represent their fluctuating components, then, $\rho$, $u$, and $p$ can be expressed as
\begin{subequations}
    \begin{align}
        \rho(x,t) &= \overline{\rho}(x)+\rho'(x,t), \\
        p(x,t) &= \overline{p}(x)+p'(x,t), \\
        u(x,t) &= \overline{u}(x)+u'(x,t).
    \end{align}
\end{subequations}
Here, $ \overline{\rho}$, $\overline{u}$, and $\overline{p}$ are the steady-state components of $\rho$, $u$, and $p$, respectively. Substituting the above expressions in Eqs.~(\ref{Eq:1}) and (\ref{Eq:2}), and subtracting the mean components produce the linearized continuity and momentum equations as follows:
\begin{align}
    \frac{\partial \rho'}{\partial t}+\overline{u}\frac{\partial\rho'}{\partial x}+u'\frac{\partial\overline{\rho}}{\partial x}+\overline{\rho}\frac{\partial u'}{\partial x}+\rho'\frac{\partial\overline{u}}{\partial x} &= 0, \label{Eq:4} \\ 
    \overline{\rho}\frac{\partial u'}{\partial t}+\rho'\overline{u}\frac{\partial\overline{u}}{\partial x}+\overline{\rho}u'\frac{\partial\overline{u}}{\partial x}+\overline{\rho}\,\overline{u}\frac{\partial u'}{\partial x}+\frac{\partial p'}{\partial x} &= 0. 
\end{align}
The assumptions that the fluid medium is a perfect gas and the propagation of sound inside the duct is an isentropic process yields\cite{Munjal2014}
\begin{align}
    \left(\frac{\partial p'}{\partial \rho'}\right)_s &= \overline{c}^2,  \label{Eq:6} \\
    \frac{\partial\overline{\rho}}{\partial x} &= \frac{\overline{\rho}}{\gamma\overline{p}} \frac{\partial\overline{p}}{\partial x}, \label{Eq:7}
\end{align}
where $\overline{c}=\sqrt{\gamma R\overline{T}}$ is the speed of sound, $\gamma$ is the specific heat ratio, $R$ is the universal gas constant, and $\overline{T}$ is the steady-state mean temperature. The subscript $s$ denotes the is isentropic process. Refer to Appendix \ref{Append:A} for the derivation. Upon using Eqs.~(\ref{Eq:6}) and (\ref{Eq:7}), and assuming that the time dependency is harmonic in nature, that is, $\rho'(x,t)=\widehat{\rho}(x)\,e^{-j\omega t}$, $p'(x,t)=\widehat{p}(x)\,e^{-j\omega t}$, and $u'(x,t)=\widehat{u}(x)\,e^{-j\omega t}$, where $\omega$ is the angular frequency, the continuity and momentum equation can be written as
\begin{align}
    \left(j\omega+\gamma\frac{d\overline{u}}{dx}\right)\widehat{p}+\overline{u}\frac{d\widehat{p}}{dx}+\frac{d\overline{p}}{dx}\widehat{u}+\gamma\overline{p}\frac{d\widehat{u}}{dx} &= 0, \label{Eq:9}\\
    \left(j\omega+\frac{d\overline{u}}{dx}\right)\widehat{u}+\overline{u}\frac{d\widehat{u}}{dx}+\frac{\overline{u}}{\gamma\overline{p}}\frac{d\overline{u}}{dx}\widehat{p}+\frac{1}{\overline{\rho}}\frac{d\widehat{p}}{dx} &= 0. \label{Eq:10}
\end{align}
The steady-state continuity and momentum equations gives the relations between $\overline{\rho}$, $\overline{p}$, and $\overline{u}$ as follows
\begin{align}
    \frac{d\overline{u}}{dx} &= -\overline{u}\left(\frac{1}{\overline{\rho}}\frac{d\overline{\rho}}{dx}\right), \label{Eq:11}\\
    \frac{d\overline{p}}{dx} &= -\overline{\rho}\, \overline{u}\frac{d\overline{u}}{dx}. \label{Eq:12}
\end{align}
If 
\begin{equation}
    \alpha = \frac{1}{\overline{\rho}}\frac{d\overline{\rho}}{dx},
\end{equation}
then Eqs.~(\ref{Eq:11}) and (\ref{Eq:12}) can be written as 
\begin{align}
    \frac{d\overline{u}}{dx} &= -\overline{u}\alpha, \label{Eq:14}\\
    \frac{d\overline{p}}{dx} &= \overline{\rho}\, \overline{u}^2\alpha. \label{Eq:15}
\end{align}
Now, the term $\widehat{u}\,d\overline{p}/dx$ in the continuity equation (Eq.~(\ref{Eq:9})) can be evaluated from the momentum equation (Eq.~(\ref{Eq:10})) and Eq.~(\ref{Eq:15}) as follows\cite{Li2017}
\begin{equation}
    \frac{d\overline{p}}{dx}\widehat{u} = -\overline{\rho}\, \overline{u}^2\alpha \times \frac{1}{\chi}\left(\overline{u}\frac{d\widehat{u}}{dx}+\frac{\overline{u}}{\gamma\overline{p}}\frac{d\overline{u}}{dx}\widehat{p}+\frac{1}{\overline{\rho}}\frac{d\widehat{p}}{dx}\right), \label{Eq:16}
\end{equation}
where $\chi = j\omega+d\overline{u}/dx$. Using Eq.~(\ref{Eq:14}), $1/\chi$ can be simplified as
\begin{equation}
    \frac{1}{\chi} = \frac{1}{j\omega-\overline{u}\alpha} = \frac{1}{jk\overline{c}}\left(1+j\frac{M\alpha}{k}\right)^{-1}, \label{Eq:17}
\end{equation}
where $k=\omega/\overline{c}$ is the wavenumber, and $M=\overline{u}/\overline{c}$ is the mean flow Mach number. For $|M\alpha|<<k$, Eq.~(\ref{Eq:17}) can be written as
\begin{equation}
    \frac{1}{\chi} = \frac{1}{jk\overline{c}}\left(1-j\frac{M\alpha}{k}\right).
\end{equation}
Now, Eq.~(\ref{Eq:16}) can be written as
\begin{equation}
    \frac{d\overline{p}}{dx}\widehat{u} = -\left(\frac{M\alpha}{jk}-\frac{M^2\alpha^2}{k^2}\right) \left(\gamma\overline{p}M^2\frac{d\widehat{u}}{dx}+M^2\frac{d\overline{u}}{dx}\widehat{p}+\overline{u}\frac{d\widehat{p}}{dx}\right).
\end{equation}
Neglecting terms that contain the Mach number order beyond $M^2$ yields
\begin{equation}
    \frac{d\overline{p}}{dx}\widehat{u} = -\left(\frac{M\alpha}{jk}-\frac{M^2\alpha^2}{k^2}\right)\overline{u}\frac{d\widehat{p}}{dx}.
\end{equation}
Substituting it in Eq.~(\ref{Eq:9}) and normalizing with respect to $\gamma\overline{p}$ gives
\begin{equation}
    \mathcal{A}\,\widehat{p}+\mathcal{B}\frac{d\widehat{p}}{dx}+\frac{d\widehat{u}}{dx} = 0, \label{Eq:21}
\end{equation}
where
\begin{align}
    \mathcal{A} &= \frac{1}{\gamma\overline{p}}\left(j\omega+\gamma\frac{d\overline{u}}{dx}\right) = \frac{1}{\gamma\overline{p}}\left(j\omega-\gamma\,\overline{u}\,\alpha\right) , \\
    \mathcal{B} &= \frac{\overline{u}}{\gamma\overline{p}}\left(1-\frac{M\alpha}{jk}+\frac{M^2\alpha^2}{k^2}\right) = \frac{M^2}{\overline{\rho}\,\overline{u}}\left(1-\frac{M\alpha}{jk}+\frac{M^2\alpha^2}{k^2}\right).
\end{align}
Similarly, normalizing the momentum equation with respect to $\overline{u}$ gives
\begin{equation}
    \mathcal{C}\,\widehat{u}+\mathcal{D}\frac{d\widehat{p}}{dx}+\frac{d\widehat{u}}{dx}+\mathcal{F}\,\widehat{p} = 0,  \label{Eq:24}
\end{equation}
where
\begin{align}
    \mathcal{C} &= \frac{1}{\overline{u}}\left(j\omega+\frac{d\overline{u}}{dx}\right) = \frac{j\omega}{\overline{u}}-\alpha, \\
    \mathcal{D} &= \frac{1}{\overline{\rho}\,\overline{u}}, \\
    \mathcal{F} &= \frac{1}{\gamma\overline{p}}\frac{d\overline{u}}{dx} = -\frac{\overline{u}\alpha}{\gamma\overline{p}} = -\frac{M^2\alpha}{\overline{\rho}\,\overline{u}}.
\end{align}
Eliminating $\widehat{u}$ terms from Eqs.~(\ref{Eq:21}) and (\ref{Eq:24}) gives
\begin{equation}
    \zeta_1\frac{d^2\widehat{p}}{dx^2}+\zeta_2\frac{d\widehat{p}}{dx}+\zeta_3\widehat{p} = 0, \label{Eq:28}
\end{equation}
where
\begin{align}
    \zeta_1 &= 1-M^2+j\frac{2M^2}{k}\frac{dM}{dx}, \\
    \zeta_2 &= \left(1-(3+\gamma)M^2\right)\alpha+j\left(2Mk+\frac{M\beta}{k}-\frac{2M\alpha^2}{k}\right), \\
    \zeta_3 &= k^2+(2-\gamma)M^2\beta+(4\gamma-5)M^2\alpha^2+j\left((2+\gamma)Mk\alpha-2\gamma kM^2\frac{dM}{dx}\right), \\
    \beta &= \frac{1}{\overline{\rho}}\frac{d^2\overline{\rho}}{dx^2}.
\end{align}
These coefficients can be calculated for a given temperature profile and mean inlet conditions of the duct. In order to do that, the mean flow variables $\overline{\rho}(x)$, $\overline{p}(x)$, and $\overline{u}(x)$ have to be evaluated from the fundamental of the fluid dynamics and thermodynamic relations.

Consider the momentum equation with steady-state mean flow variables, that is, 
\begin{equation}
    \overline{\rho}\,\overline{u}\frac{d\overline{u}}{dx}+\frac{d\overline{p}}{dx} = 0.
\end{equation}
Assuming the duct inlet starts at $x=$ 0, using the first-order approximation of the Maclaurin series\cite{Beyer1987} (Taylor series expansion of a function about 0), the above momentum equation can be written as
\begin{equation}
    \overline{\rho}_0\overline{u}_0(\overline{u}(x)-\overline{u}_0)+\overline{p}(x)-\overline{p}_0 = 0, \label{Eq:34}
\end{equation}
where $\overline{\rho}_0$, $\overline{p}_0$, and $\overline{u}_0$ are the steady-state mean density, pressure, and velocity at the inlet, that is, at $x=$ 0. From the perfect gas law, it can be written that
\begin{equation}
    \frac{\overline{p}(x)}{\overline{\rho}(x)\overline{T}(x)} = \frac{\overline{p}_0}{\overline{\rho}_0\overline{T}_0},
\end{equation}
where $\overline{T}_0$ is the mean inlet temperature. Now, using the mass continuity in a uniform duct, that is, $\overline{\rho}\,\overline{u}=\overline{\rho}_0\overline{u}_0$, $\overline{p}(x)$ can be written as
\begin{equation}
    \overline{p}(x) = \frac{\overline{u}_0\overline{T}(x)}{\overline{u}(x)\overline{T}_0}.
\end{equation}
Substituting it in Eq.~(\ref{Eq:34}) gives a quadratic equation in $\overline{u}(x)$ as follows\cite{Li2017}
\begin{equation}
    \overline{\rho}_0\overline{u}_0\overline{u}^2(x)-(\overline{p}_0+\overline{\rho}_0\overline{u}^2_0)\overline{u}(x)+\overline{p}_0\overline{u}_0\frac{\overline{T}(x)}{\overline{T}_0} = 0. \label{Eq:37}
\end{equation}
It will have two roots. Usually, the smallest root of the two should be considered, since the other root violates the conservation laws. 

Now, the mean pressure $\overline{p}(x)$ can be calculated from the mean velocity $\overline{u}(x)$ using Eq.~(\ref{Eq:34}) as follows
\begin{equation}
    \overline{p}(x) = \overline{p}_0+\overline{\rho}_0\overline{u}_0(\overline{u}_0-\overline{u}(x)), \label{Eq:38}
\end{equation}
and the mean density $\overline{\rho}(x)$ can be calculated from the perfect gas law as follows
\begin{equation}
    \overline{\rho}(x) = \frac{\overline{p}(x)}{R\overline{T}(x)}.
\end{equation}
The other parameters: $k$, $M$, $dM/dx$, $\alpha$, and $\beta$ can be calculated from $\overline{\rho}$, $\overline{p}$, and $\overline{u}$. Refer to Appendix B for more details.

\section{Deep neural network formulation} \label{Sec:3}
\begin{figure}[h!]
\includegraphics[scale=0.8]{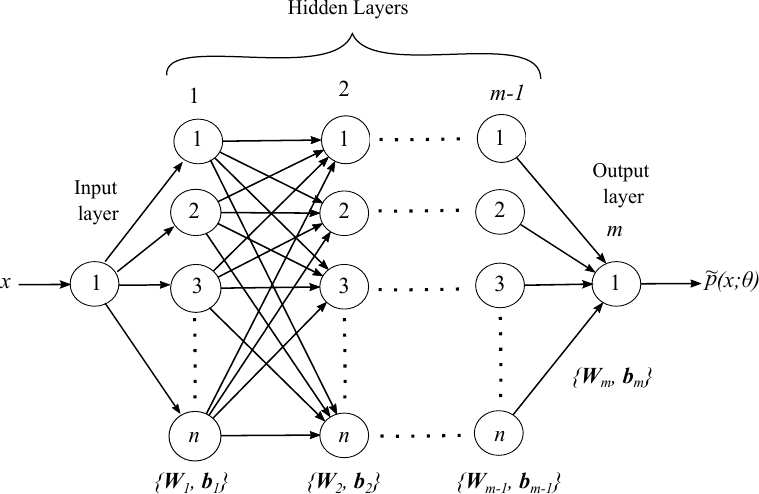}\centering
\caption{\label{fig:1}{Schematic diagram of a feedforward neural network.}}
\end{figure}

According to the universal approximation theorem\cite{Hornik1989}, the acoustic pressure $\widehat{p}$ in Eq.~(\ref{Eq:28}) can be approximated to the output of a feedforward neural network shown in Fig.~\ref{fig:1}. The neural network takes the domain information in the discretized format at the input-layer. It will undergo a nonlinear function known as the activation function ($\sigma$) at each neuron in the hidden-layers. The required output is returned at the output-layer through a linear activation function. 

If $\mathbf{f}_0$, $\mathbf{f}_q$, and $\mathbf{f}_m$ represent the output of the input, hidden, and output-layers, respectively, then, the feedforward neural network shown in Fig.~1 can be expressed in the mathematical form as follows\cite{Dung2023}
\begin{align}
    \mathbf{f}_0 &= \mathbf{x}, \\
    \mathbf{f}_q &= \sigma(\mathbf{W}_q\mathbf{f}_{q-1}+\mathbf{b}_q), \\
    \mathbf{f}_m &= \mathbf{W}_m\mathbf{f}_{m-1}+\mathbf{b}_m,
\end{align}
where $q=$ 1, 2, 3, ..., $m-1$. The output of the neural network $\mathbf{f}_m$, which is represented by $\widetilde{p}(x;\theta)$, is an approximation to the acoustic pressure $\widehat{p}(x)$. Here,  $\theta=\left\{\mathbf{W}_q, \mathbf{b}_q, \mathbf{W}_m, \mathbf{b}_m\right\}$ are the parameters of the network. These can be found through an optimization procedure.

For the duct of length $L$, the optimization problem can be formulated as\cite{Basir2022}
\begin{equation}
\begin{aligned}
\min_{\theta} \quad & \mathcal{L}_d(\mathbf{x}_d;\theta), \quad \mathbf{x}_d\in(0,L) \\
\textrm{s.t.} \quad & \mathcal{L}_b(\mathbf{x}_b;\theta)=0, \quad \mathbf{x}_b\in\{0,L\} \label{Eq:43}
\end{aligned}
\end{equation}
where $\mathcal{L}_d$ and $\mathcal{L}_b$ are the loss functions associated with the differential equation and the boundary conditions, respectively. It is a constrained optimization problem. It can be converted into an unconstrained optimization problem using the trial solution method\cite{Lagaris1998}. According to this method, a trial solution $\widetilde{p}_t(x;\theta)$ is constructed in such a way that it always satisfies the given boundary conditions prior to the training process, as follows
\begin{equation}
    \widetilde{p}_t(x;\theta) = \frac{L-x}{L}\widehat{p}_0+\frac{x}{L}\widehat{p}_L+\frac{x(L-x)}{L^2}\widetilde{p}(x;\theta),
\end{equation}
where $\widehat{p}_0$ and $\widehat{p}_L$ are the boundary values at $x=0$ and $x=L$, respectively. It can be observed that the first two terms ensure that $\widetilde{p}_t(x;\theta)$ satisfies the prescribed boundary conditions prior to the training process. Using the trial solution method, the optimization problem in Eq.~(\ref{Eq:43}) can be posed as an unconstrained optimization problem as follows
\begin{equation}
\min_{\theta} \quad  \mathcal{L}_d(\mathbf{x};\theta), \quad \mathbf{x}\in[0,L].  \label{Eq:45}
\end{equation}
Here, the loss function $\mathcal{L}_d$ is evaluated from Eq.~(\ref{Eq:28}). 

It is known that to perform an optimization procedure, the loss function needs to be a real-valued function. However, the presence of complex-valued coefficients in Eq.~(\ref{Eq:28}) makes the acoustic pressure a complex-valued function. Subsequently, the loss function will become a complex-valued function. This problem can be bypassed by splitting the Eq.~(\ref{Eq:28}) into real and imaginary parts\cite{Raissi2019}.

Let 
\begin{subequations}
    \begin{align}
        \zeta_1 &= \zeta_1^R+j\zeta_1^I, \\
        \zeta_2 &= \zeta_2^R+j\zeta_2^I, \\
        \zeta_3 &= \zeta_3^R+j\zeta_3^I, 
\end{align}
\end{subequations}
be the complex-valued coefficients corresponding to the complex-valued acoustic pressure $\widehat{p} = \widehat{p}^R+j\widehat{p}^I$ at a given frequency. Here, the superscripts $R$ and $I$ denote the real and imaginary-parts, respectively. Upon substituting these expressions in Eq.~(\ref{Eq:28}) gives two governing equations, one associated with the real part and the other associated with the imaginary-part as follows \\
\underline{Governing equation associated with the real-part:} 
\begin{equation}
    \zeta_1^R\frac{d^2\widehat{p}^R}{dx^2}-\zeta_1^I\frac{d^2\widehat{p}^I}{dx^2}+\zeta_2^R\frac{d\widehat{p}^R}{dx}-\zeta_2^I\frac{d\widehat{p}^I}{dx}+\zeta_3^R\widehat{p}^R-\zeta_3^I\widehat{p}^I = 0.
\end{equation}
\underline{Governing equation associated with the imaginary-part:} 
\begin{equation}
    \zeta_1^R\frac{d^2\widehat{p}^I}{dx^2}+\zeta_1^I\frac{d^2\widehat{p}^R}{dx^2}+\zeta_2^R\frac{d\widehat{p}^I}{dx}+\zeta_2^I\frac{d\widehat{p}^R}{dx}+\zeta_3^R\widehat{p}^I+\zeta_3^I\widehat{p}^R = 0.
\end{equation}

\begin{figure}[h!]
\includegraphics[scale=0.8]{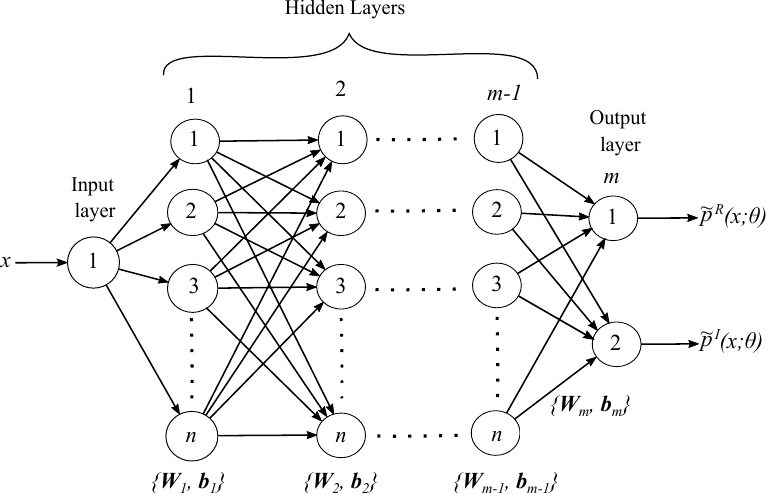}\centering
\caption{\label{fig:2}{Schematic diagram of a feedforward neural network for the complex-valued acoustic pressure.}}
\end{figure}

Now, the loss function in Eq.~(\ref{Eq:45}) will have two terms as follows
\begin{equation}
    \mathcal{L}_d = \mathcal{L}_d^R+\mathcal{L}_d^I,
\end{equation}
where $\mathcal{L}_d^R$ and $\mathcal{L}_d^I$ are the loss functions associated with the real and imaginary-parts of the governing equation, respectively. These can be calculated as
\begin{multline}
    \mathcal{L}_d^R = \frac{1}{N}\sum_{i=1}^N\left\|\zeta_1^R\left.\left(\frac{d^2}{dx^2}\widetilde{p}_t^R(x;\theta)\right)\right|_{x=x^{(i)}}-\zeta_1^I\left.\left(\frac{d^2}{dx^2}\widetilde{p}_t^I(x;\theta)\right)\right|_{x=x^{(i)}}\right. \\
    +\left.\zeta_2^R\left.\left(\frac{d}{dx}\widetilde{p}_t^R(x;\theta)\right)\right|_{x=x^{(i)}}-\zeta_2^I\left.\left(\frac{d}{dx}\widetilde{p}_t^I(x;\theta)\right)\right|_{x=x^{(i)}}+\zeta_3^R\,\widehat{p}^R(x^{(i)};\theta)-\zeta_3^I\,\widehat{p}^I(x^{(i)};\theta) \right\|^2_2,
\end{multline}
    
\begin{multline}
    \mathcal{L}_d^I = \frac{1}{N}\sum_{i=1}^N\left\|\zeta_1^R\left.\left(\frac{d^2}{dx^2}\widetilde{p}_t^I(x;\theta)\right)\right|_{x=x^{(i)}}+\zeta_1^I\left.\left(\frac{d^2}{dx^2}\widetilde{p}_t^R(x;\theta)\right)\right|_{x=x^{(i)}}\right. \\
    +\left.\zeta_2^R\left.\left(\frac{d}{dx}\widetilde{p}_t^I(x;\theta)\right)\right|_{x=x^{(i)}}+\zeta_2^I\left.\left(\frac{d}{dx}\widetilde{p}_t^R(x;\theta)\right)\right|_{x=x^{(i)}}+\zeta_3^R\,\widehat{p}^I(x^{(i)};\theta)+\zeta_3^I\,\widehat{p}^R(x^{(i)};\theta) \right\|^2_2.
\end{multline}
Here, $\left\|\,\cdot\,\right\|_2$ denotes the $L_2$-norm, $N$ represents the total number of collocation points with the $i$-th point represented by $x^{(i)}$. $\widetilde{p}_t^R$ and $\widetilde{p}_t^I$ are the trial solutions associated with the real and imaginary parts of the governing equation, respectively. If $\widehat{p}_0=\widehat{p}^R_0+j\widehat{p}^I_0$ and $\widehat{p}_L=\widehat{p}^R_L+j\widehat{p}^I_L$, then, the trial solutions $\widetilde{p}_t^R$ and $\widetilde{p}_t^I$ can be constructed as follows
\begin{align}
    \widetilde{p}_t^R(x;\theta) &= \frac{L-x}{L}\widehat{p}_0^R+\frac{x}{L}\widehat{p}_L^R+\frac{x}{L}\frac{L-x}{L}\widetilde{p}^R(x;\theta), \\
    \widetilde{p}_t^I(x;\theta) &= \frac{L-x}{L}\widehat{p}_0^I+\frac{x}{L}\widehat{p}_L^I+\frac{x}{L}\frac{L-x}{L}\widetilde{p}^I(x;\theta). 
\end{align}
To predict complex-valued acoustic pressure, that is, $\widetilde{p}_t=\widetilde{p}_t^R+j\widetilde{p}_t^I$, the feedforward neural network shown in Fig.~\ref{fig:1} cannot be used as it has only one neuron in the output-layer. The output layer must be modified in such a way that it has two neurons; one is to predict the real-part of the acoustic pressure $\widetilde{p}^R$ and the other is to predict the imaginary-part of the acoustic pressure $\widetilde{p}^I$ as shown in Fig.~\ref{fig:2}. 

\section{Results and discussion} \label{Sec:4}
\begin{figure}[h!]
\includegraphics[scale=0.8]{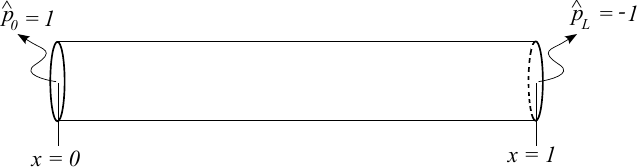}\centering
\caption{\label{fig:3}{Schematic diagram of a uniform duct with boundary conditions.}}
\end{figure}

To predict complex-valued acoustic pressure, a uniform duct of length $L=$ 1 m is considered, as shown in Fig.~\ref{fig:3}. The boundary conditions of the duct $x=$ 0, and $x=$ 1 are assumed to be $\widehat{p}_0=$ 1 Pa, and $\widehat{p}_L=$ -1 Pa, respectively. The medium inside the duct is assumed to be air. The steady-state mean properties of air at the inlet and temperature at the outlet are considered along similar lines to those existing in the literature\cite{Li2017} and are given in Table~\ref{tab:1}.

\begin{table}[th]
\caption{Steady-state mean properties of the air at the inlet along with the outlet temperature.} \label{tab:1}
\centering
{\begin{tabular}{@{}cccccc@{}} \toprule
$\overline{p}_0$ & $\overline{T}_0$ & $\overline{T}_L$  & $M_0$  & $\gamma$ & $R$\\
(Pa) & (K) & (K) & (-) & (-) & (J/kg.K)  \\ \colrule
1$\times$10$^{5}$ & 1600 & 800 & 0.2 & 1.4 & 287 \\ \botrule
\end{tabular}}
\end{table}
 
\begin{figure}[h!]
\includegraphics[scale=0.9]{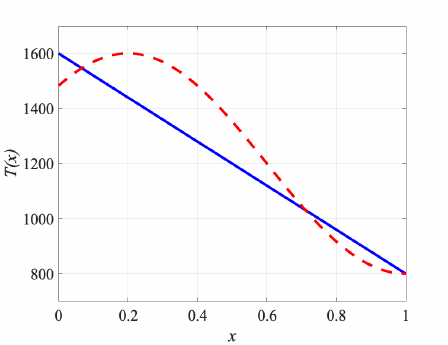}\centering
\caption{\label{fig:4}{Temperature profiles:\protect\blueline Linear temperature profile, \protect\redline Sinusoidal temperature profile.}}
\end{figure}

Two temperature profiles, namely, linear and sinusoidal, as given in Eqs.~(\ref{Eq:54}) and (\ref{Eq:55}), respectively, are considered for analysis purposes. 
\begin{align}
    \overline{T}(x) &= \overline{T}_0+\overline{T}_mx, \label{Eq:54} \\
    \overline{T}(x) &= \frac{1}{2}\left[\overline{T}_d\sin\left(\frac{5\pi}{4}\frac{x}{L}+\frac{\pi}{4}\right)+\overline{T}_s\right], \label{Eq:55}
\end{align}
where
\begin{subequations}
    \begin{align}
        \overline{T}_d &= \overline{T}_0-\overline{T}_L, \\  
        \overline{T}_s &= \overline{T}_0+\overline{T}_L, \\ 
        \overline{T}_m &= -\frac{\overline{T}_d}{L}.
    \end{align}
\end{subequations}
Here, $\overline{T}_L$ is the mean steady-state temperature at the outlet. The variation in temperature with respect to position in both profiles is shown in Fig.~\ref{fig:4}. It can be observed that the sinusoidal temperature profile is constructed in such a way that the maximum temperature ($T_0$) occurs inside the duct rather than at the inlet boundary, which is the case in the linear temperature profile. 

To perform the optimization procedure, a feedforward neural network is constructed with seven layers ($m=$ 7), and 90 neurons ($n=$ 90) in each hidden-layer. The domain of length 1 m is divided into 10000 random collocation points ($N=$ 10000). The network is initialized with the He initialization\cite{He2015}. The optimization is carried out using the L-BFGS optimizer and \emph{sine} activation function. The formulation is implemented in MATLAB (Version 2023a) using the Deep Learning Toolbox$^{\text{\texttrademark}}$ and the Statistics and Machine Learning Toolbox$^{\text{\texttrademark}}$.

\subsection{Prediction of acoustic pressure}
\begin{figure}[h!]
\includegraphics[scale=0.9]{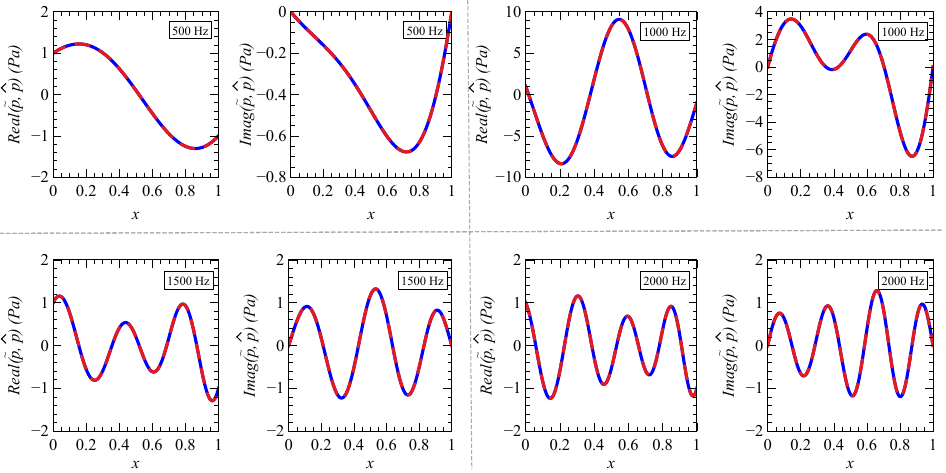}\centering
\caption{\label{fig:5}{Acoustic pressure with linear temperature profile:\protect\blueline True solution, \protect\redline Predicted solution.}}
\end{figure}

The analysis is carried out up to 2000 Hz, starting from 500 Hz, with increments of 500 Hz. Fig.~\ref{fig:5} shows the comparison of the acoustic pressure obtained from the neural network formulation (predicted solution) against that obtained from the traditional Runge-Kutta solver \texttt{bvp4c} in MATLAB (true solution) for the linear temperature profile. It can be observed that the predicted solution is in good agreement with the true solution at all the frequencies considered. 

The relative error between the two methods is calculated as 
\begin{equation}
    \delta p = \frac{\sqrt{\displaystyle\sum_{i=1}^{N_t} |\widetilde{p}_t(x^{(i)};\theta)-\widehat{p}(x^{(i)})|^2}}{\sqrt{\displaystyle\sum_{i=1}^{N_t}|\widehat{p}(x^{(i)})|^2}},
\end{equation}
where $\widetilde{p}_t$ is the predicted solution, $\widehat{p}$ is the true solution, and $N_t$ represents the number of test points. A total of 500 linearly spaced test points have been used in each frequency comparison graph ($N_t=$ 500), and the same has been used for the calculation of $\delta p$. The relative errors are calculated for the real and imaginary components separately and are tabulated at the individual frequencies in Table~\ref{tab:2}. It can be observed that the error between the two methods is insignificant.
\begin{table}[th]
\caption{Relative error between the predicted solution and true solution with linear temperature profile.} \label{tab:2}
\centering
{\begin{tabular}{@{}ccc@{}} \toprule
Frequency & $\delta p$ & $\delta p$ \\
(Hz) & (Real-part) & (Imaginary-part)  \\ \colrule
500 & 4.80$\times$10$^{-7}$ & 8.05$\times$10$^{-7}$ \\ 
1000 & 3.82$\times$10$^{-5}$ & 7.51$\times$10$^{-5}$ \\
1500 & 5.24$\times$10$^{-6}$ & 7.94$\times$10$^{-6}$ \\
2000 & 4.23$\times$10$^{-5}$ & 4.38$\times$10$^{-5}$ \\
\botrule
\end{tabular}}
\end{table}

\begin{figure}[h!]
\includegraphics[scale=0.9]{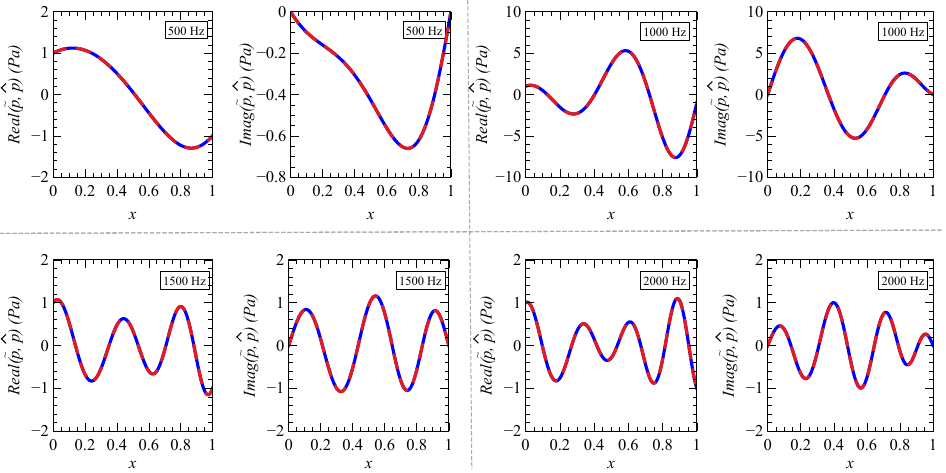}\centering
\caption{\label{fig:6}{Acoustic pressure with sinusoidal temperature profile:\protect\blueline True solution, \protect\redline Predicted solution.}}
\end{figure}

Similar observations can also be made in the case of sinusoidal temperature variation. Fig.~\ref{fig:6} shows the comparison of the acoustic pressure with the two methods for the sinusoidal temperature gradient. The corresponding relative errors are mentioned in Table~\ref{tab:3}. From the results, it can be understood that the proposed neural network formulation is able to capture variations in the acoustic pressures with the sinusoidal temperature gradients successfully. Note here that the same neural network architecture and discretization have been used for both temperature profiles. 

\begin{table}[th]
\caption{Relative error between the predicted solution and true solution with sinusoidal temperature profile.} \label{tab:3}
\centering
{\begin{tabular}{@{}ccc@{}} \toprule
Frequency & $\delta p$ & $\delta p$ \\
(Hz) & (Real-part) & (Imaginary-part)  \\ \colrule
500 & 3.10$\times$10$^{-7}$ & 4.42$\times$10$^{-7}$ \\ 
1000 & 1.92$\times$10$^{-5}$ & 2.56$\times$10$^{-5}$ \\
1500 & 1.07$\times$10$^{-6}$ & 1.50$\times$10$^{-6}$ \\
2000 & 3.40$\times$10$^{-5}$ & 3.80$\times$10$^{-5}$ \\
\botrule
\end{tabular}}
\end{table}

\subsection{Prediction of particle velocity}
\begin{figure}[h!]
\includegraphics[scale=0.85]{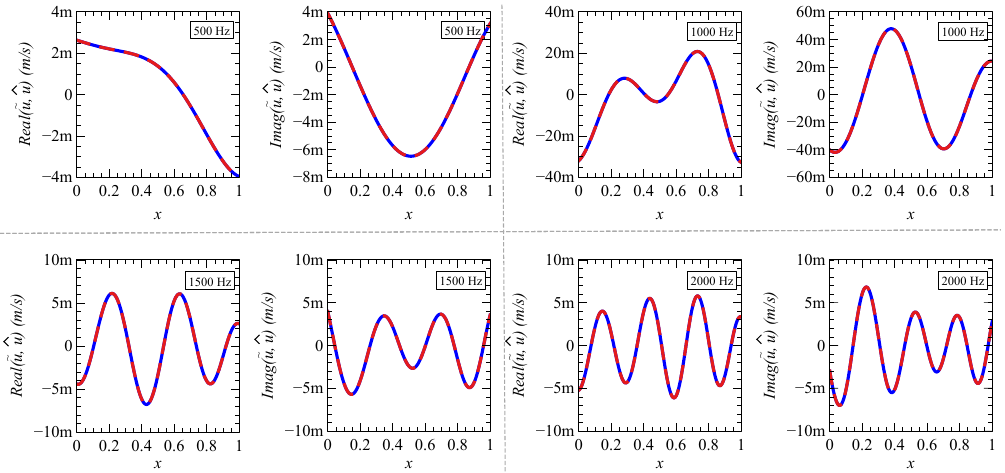}\centering
\caption{\label{fig:7}{Particle velocity with linear temperature profile:\protect\blueline True solution, \protect\redline Predicted solution.}}
\end{figure}

It is essential to predict the particle velocity besides the acoustic pressure to calculate quantities such as acoustic intensity, power, impedance, transfer functions, etc. The particle velocity can be calculated from the predicted acoustic pressure using the continuity and momentum equation\cite{Munjal2014,Li2017}. Consider Eqs.~(\ref{Eq:21}) and (\ref{Eq:24}). Eliminating $d\widehat{u}/dx$ from both equations gives the relation
\begin{equation}
    \widehat{u} = \left(\frac{\mathcal{A}-\mathcal{F}}{\mathcal{C}}\right)\widehat{p}+\left(\frac{\mathcal{B}-\mathcal{D}}{\mathcal{C}}\right)\frac{d\widehat{p}}{dx}. \label{Eq:58}
\end{equation}
Using this relation, the particle velocity can be predicted through the neural network formulation as
\begin{equation}
    \widetilde{u}_t(x^{(i)};\theta) = \left(\frac{\mathcal{A}-\mathcal{F}}{\mathcal{C}}\right)\widetilde{p}_t(x^{(i)};\theta)+\left(\frac{\mathcal{B}-\mathcal{D}}{\mathcal{C}}\right)\frac{d}{dx^{(i)}}\widetilde{p}_t(x^{(i)};\theta),
\end{equation}
where $i=$ 1, 2, 3, ..., $N$. The gradient of acoustic pressure with respect to $x$ can be evaluated using the automatic differential algorithm\cite{Lagaris1998}. The real and imaginary parts of the particle velocity thus obtained for the linear temperature profile at different frequencies are shown in Fig.~\ref{fig:7} against the true solution. Here, the true solution is obtained using the \texttt{bvp4c} solver in conjunction with Eq.~\ref{Eq:58}. In the graphs, $m$ in the ordinate values indicate milli units. The relative error between the two solutions is calculated and is tabulated in Table~\ref{tab:4}. The results indicate that the neural network formulation is able to capture the tiny particle velocity fluctuations with great accuracy.

\begin{table}[th]
\caption{Relative error between the predicted solution and true solution with linear temperature profile.} \label{tab:4}
\centering
{\begin{tabular}{@{}ccc@{}} \toprule
Frequency & $\delta u$ & $\delta u$ \\
(Hz) & (Real-part) & (Imaginary-part)  \\ \colrule
500 & 6.99$\times$10$^{-6}$ & 4.26$\times$10$^{-6}$ \\ 
1000 & 8.32$\times$10$^{-5}$ & 3.49$\times$10$^{-5}$ \\
1500 & 8.81$\times$10$^{-6}$ & 9.34$\times$10$^{-6}$ \\
2000 & 4.52$\times$10$^{-5}$ & 4.29$\times$10$^{-5}$ \\
\botrule
\end{tabular}}
\end{table}

\begin{figure}[h!]
\includegraphics[scale=0.85]{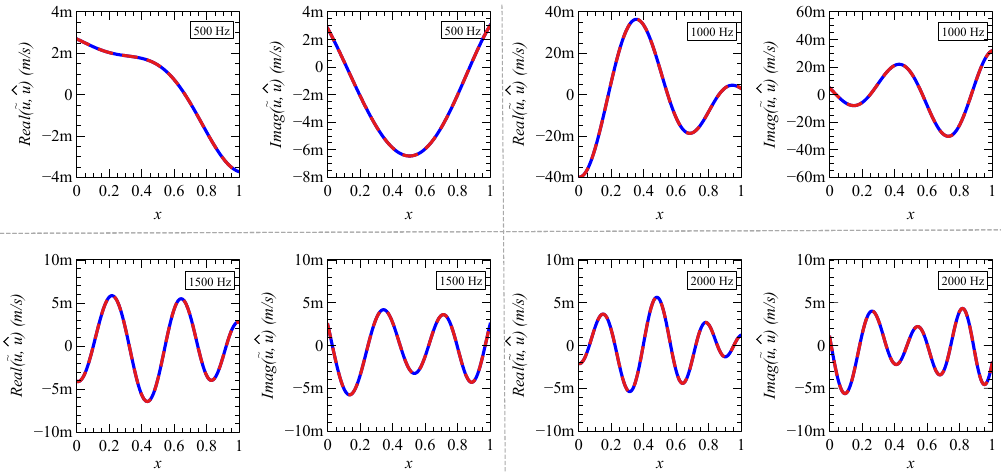}\centering
\caption{\label{fig:8}{Particle velocity with sinusoidal temperature profile:\protect\blueline True solution, \protect\redline Predicted solution.}}
\end{figure}

Similar results can also be observed with the sinusoidal temperature profile in Fig.~\ref{fig:8} and in Table~\ref{tab:5}. 

\begin{table}[th]
\caption{Relative error between the predicted solution and true solution with sinusoidal temperature profile.} \label{tab:5}
\centering
{\begin{tabular}{@{}ccc@{}} \toprule
Frequency & $\delta u$ & $\delta u$ \\
(Hz) & (Real-part) & (Imaginary-part)  \\ \colrule
500 & 4.19$\times$10$^{-6}$ & 3.24$\times$10$^{-6}$ \\ 
1000 & 2.57$\times$10$^{-5}$ & 2.06$\times$10$^{-5}$ \\
1500 & 7.79$\times$10$^{-6}$ & 4.71$\times$10$^{-6}$ \\
2000 & 4.01$\times$10$^{-5}$ & 3.49$\times$10$^{-5}$ \\
\botrule
\end{tabular}}
\end{table}

It should be noted here that the particle velocity is evaluated from the predicted acoustic field using the continuity and momentum equations. Alternately, one can predict the particle velocity from the momentum equation alone using the transfer learning technique available in machine learning methods\cite{Xu2023,Gao2022,Pellegrin2022,Desai2021}. According to this method, a separate feedforward neural network $\widetilde{u}_t(x;\tilde{\theta})$ that approximates the particle velocity $\widehat{u}$ will be constructed. The parameters of the network $\tilde{\theta}$ are found by solving the following optimization problem
\begin{equation}
\min_{\tilde{\theta}} \quad  \mathcal{L}_u^R(\mathbf{x};\tilde{\theta})+\mathcal{L}_u^I(\mathbf{x};\tilde{\theta}),
\end{equation}
where $\mathcal{L}_u^R$ and $\mathcal{L}_u^I$ are the loss functions associated with the real and imaginary-parts of the momentum equation, respectively. These can be calculated from Eq.~(\ref{Eq:24}) as follows
\begin{multline}
    \mathcal{L}_u^R(x;\tilde{\theta})=\frac{1}{N_u}\sum_{i=1}^{N_u}\left\|\mathcal{C}^R\,\widetilde{u}_t^R(x^{(i)};\tilde{\theta})-\mathcal{C}^I\,\widetilde{u}_t^I(x^{(i)};\tilde{\theta})+\mathcal{D}^R\left.\left(\frac{d}{dx}\widetilde{p}_t^R(x;\theta)\right)\right|_{x=x^{(i)}}\right. \\ 
    \left.-\mathcal{D}^I\left.\left(\frac{d}{dx}\widetilde{p}_t^I(x;\theta)\right)\right|_{x=x^{(i)}}+\left.\left(\frac{d}{dx}\widetilde{u}_t^R(x;\tilde{\theta})\right)\right|_{x=x^{(i)}}+\mathcal{F}^R\,\widetilde{p}_t^R(x^{(i)};\theta)-\mathcal{F}^I\,\widetilde{p}_t^I(x^{(i)};\theta) \right\|^2_2,
\end{multline}

\begin{multline}
    \mathcal{L}_u^I(x;\tilde{\theta})=\frac{1}{N_u}\sum_{i=1}^{N_u}\left\|\mathcal{C}^R\,\widetilde{u}_t^I(x^{(i)};\tilde{\theta})+\mathcal{C}^I\,\widetilde{u}_t^R(x^{(i)};\tilde{\theta})+\mathcal{D}^R\left.\left(\frac{d}{dx}\widetilde{p}_t^I(x;\theta)\right)\right|_{x=x^{(i)}}\right. \\
    \left.+\mathcal{D}^I\left.\left(\frac{d}{dx}\widetilde{p}_t^R(x;\theta)\right)\right|_{x=x^{(i)}}+\left.\left(\frac{d}{dx}\widetilde{u}_t^I(x;\tilde{\theta})\right)\right|_{x=x^{(i)}}+\mathcal{F}^R\,\widetilde{p}_t^I(x^{(i)};\theta)+\mathcal{F}^I\,\widetilde{p}_t^R(x^{(i)};\theta) \right\|^2_2.
\end{multline}
Here, $N_u$ is the number of collation points used to approximate the particle velocity. The superscripts $R$ and $I$ of the coefficients $\mathcal{C}$, $\mathcal{D}$, and $\mathcal{F}$ denote the real and imaginary components of them, respectively.

The particle velocity $\widetilde{u}_t(x;\tilde{\theta})$ predicted using the transfer learning technique has more advantages compared to $\widetilde{u}_t(x;\theta)$ evaluated using the former method. Some of them are listed below.
\begin{arabiclist}
\item Since $\widetilde{u}_t(x;\tilde{\theta})$ is the result of a new optimization problem, the new parameters $\tilde{\theta}$ can be saved, and the model can be recalled whenever needed without explicit evaluation from the acoustic pressure field. 
\item It is not necessary to choose $N_u=N$. In other words, $\widetilde{u}_t(x;\tilde{\theta})$ can be obtained with a different set of collocation points, preferably $N_u<N$, without compromising the accuracy. 
\item It bypasses the need to perform algebraic operations that are required to obtain Eq.~(\ref{Eq:58}). This feature is useful when extending the formulation to higher dimensions.
\end{arabiclist}

In this work, the particle velocity is predicted using both methods and it is observed that both methods yield similar results. However, results from the former method are reported in Fig.~\ref{fig:7} and Fig.~\ref{fig:8} to demonstrate a few similarities of the proposed neural network formulation with the traditional analytical method. In either of the methods, it is necessary to be diligent in predicting acoustic pressure. Any error that occurred during its prediction will propagate into the prediction of the particle velocity. 

\subsection{Effect of temperature gradient}
To study the effect of the temperature gradient on the acoustic field, it is assumed that the medium inside the duct is having a uniform temperature $T_s/2$ throughout its length, that is, average of the inlet and outlet temperatures. In other words, the temperature gradient is zero, that is, $d\overline{T}/dx=$ 0. It implies that $\alpha=$ 0, $\beta=$ 0, and $dM/dx=$ 0. Substituting these parameters into Eq.~(\ref{Eq:28}) gives the governing equation with uniform mean flow as follows\cite{Munjal2014,Li2017}
\begin{equation}
    (1-M^2)\frac{d^2\widehat{p}}{dx^2}+2jkM\frac{d\widehat{p}}{dx}+k^2\widehat{p} = 0.  \label{Eq:63}
\end{equation}
This equation has been solved for $M=$ 0.2 and $\overline{c}=\sqrt{\gamma R\overline{T}_s/2}$ using the proposed neural network formulation.

\begin{figure}[h!]
\includegraphics[scale=1.1]{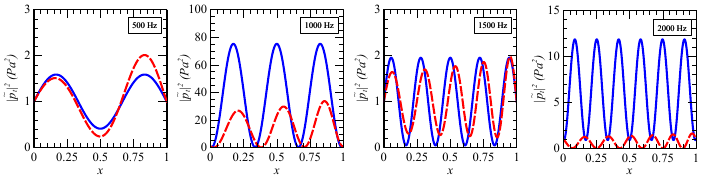}\centering
\caption{\label{fig:9}{Effect of the temperature gradient on the acoustic pressure:\protect\blueline Without temperature gradient, \protect\redline With temperature gradient (linear).}}
\end{figure}

To understand the effect of the temperature gradient, the predicted acoustic field $\widetilde{p}_t$ is multiplied by its complex conjugate $\widetilde{p}_t^{\,*}$ and the resulting acoustic amplitude\cite{Sujith1995}
\begin{equation}
    \abs{\,\widetilde{p}\,}^2 = \widetilde{p}_t\,\widetilde{p}_t^{\,*}
\end{equation}
is calculated with and without the temperature gradient. Fig.~\ref{fig:9} shows the comparison of the acoustic pressure with (linear) and without the temperature gradient at different frequencies. It can be observed that the temperature gradient has a significant effect on the acoustic field. The peak acoustic pressure amplitude is constant in the absence of the temperature gradient, whereas it increases with respect to position in the presence of the temperature gradient. This can be clearly observed at 500 Hz, 1000 Hz, and 1500 Hz. In addition, the temperature gradient reduces the peak pressure amplitudes occurring at 1000 Hz and 2000 Hz by altering the resonance frequency of the duct. These observations are along the lines of those reported in the literature\cite{Sujith1995,Karthik2000}. Similar observations can also be made with the sinusoidal temperature profile. Note here that the acoustic pressure without the temperature gradient is validated against the numerical solution of Eq.~(\ref{Eq:63}) with a maximum relative error of 0.01\%. However, these results are excluded to avoid redundancy. 

\section{Conclusion} \label{Sec:5}
Predicting acoustic field in a duct carrying a fluid whose properties change with respect to position is a challenging problem. Initially, the governing equation that represents the propagation of the sound in the duct with a heterogeneous medium is derived in the frequency domain from the fundamentals of fluid dynamics. The problem is posed as an unconstrained optimization problem solved using deep neural networks. A framework is established to predict the complex-valued acoustic field, and the acoustic pressure and particle velocity are predicted up to 2000 Hz with a maximum relative error of $\mathcal{O}(10^{-5})$. Some of the salient features of the proposed neural network formulation are as follows
\begin{arabiclist}
\item A single neural network architecture is able to predict both the acoustic pressure and particle velocity which are several order away in terms of their magnitude. 
\item The neural network model once trained can be treated as an analytical function or a pre-trained network to derive the other quantities of interest using the automatic differentiation and transfer learning techniques, respectively. 
\end{arabiclist}
These features are useful in extending the formulation to solve problems involving complicated physics, especially in higher dimensions. One limitation of PINNs is that the trained network is domain and boundary condition specific. It needs retraining for different domains and boundary conditions.  

\section*{Acknowledgments}
The authors acknowledge the support received from the Department of Science and Technology, and Science and Engineering Research Board (SERB), Government of India towards this research.

\bibliographystyle{ws-jtca} 
\bibliography{references}







\appendix

\section{Derivation of $p-\rho$ relations for an isentropic process} \label{Append:A}
Since the sound propagation inside the duct is an isentropic process, it can written that 
\begin{equation}
    \frac{p}{\rho^{\gamma}} = \text{Constant} \label{Eq:A1}
\end{equation}
Differentiating with respect to $t$ on both side gives
\begin{equation}
    \frac{\partial \rho}{\partial t} = \frac{\rho}{\gamma p} \frac{\partial p}{\partial t}
\end{equation}
Substitution of the expressions $p(x,t)=\overline{p}(x)+p'(x,t)$ and $\rho(x,t)=\overline{\rho}(x)+\rho'(x,t)$ in the above equation, and subsequent linearization gives
\begin{equation}
     \frac{\partial\rho'}{\partial t} = \frac{\overline{\rho}}{\gamma \overline{p}}\frac{\partial p'}{\partial t} \label{Eq:A3}
\end{equation}
Similarly, differentiation of Eq.~(\ref{Eq:A1}) with respect to $x$, and substitution of the expressions $p(x,t)$ and $\rho(x,t)$, followed by the linearization gives
\begin{equation}
    \gamma(\overline{p}+p')\frac{\partial\overline{\rho}}{\partial x}+\gamma\overline{p}\frac{\partial\rho'}{\partial x} = (\overline{\rho}+\rho')\frac{\partial\overline{p}}{\partial x}+\overline{\rho}\frac{\partial p'}{\partial x}.
\end{equation}
Dividing both sides by $\overline{\rho}$, and using the perfect gas law $\overline{p}=\overline{\rho}R\overline{T}$ in conjunction with the order analysis, that is, $p'<<\overline{p}$ and $\rho'<<\overline{\rho}$, gives
\begin{equation}
    \frac{\partial\overline{\rho}}{\partial x}+\frac{\partial \rho'}{\partial x} = \frac{\overline{\rho}}{\gamma \overline{p}}\left(\frac{\partial\overline{p}}{\partial x}+\frac{\partial p'}{\partial x}\right).
\end{equation}
Comparison of the mean and fluctuating components gives
\begin{align}
    \frac{\partial\overline{\rho}}{\partial x} &= \frac{\overline{\rho}}{\gamma\overline{p}} \frac{\partial\overline{p}}{\partial x}, \\
    \frac{\partial \rho'}{\partial x} &= \frac{\overline{\rho}}{\gamma\overline{p}} \frac{\partial p'}{\partial x}. \label{Eq:A7}
\end{align}
Equations~(\ref{Eq:A3}) and (\ref{Eq:A7}) together can be written as
\begin{equation}
    \left(\frac{\partial p'}{\partial \rho'}\right)_s = \overline{c}^2,
\end{equation}
where $\overline{c}=\sqrt{\gamma R\overline{T}}$ is the speed of sound. 


\section{Calculation of the parameters: $k$, $M$, $dM/dx$, $\alpha$, and $\beta$} \label{Append:B}
\subsection{Calculating $k$:}
\begin{equation}
    k=\frac{2\pi f}{\overline{c}(x)},
\end{equation}
where $f$ is the given frequency, and $\overline{c}(x)=\sqrt{\gamma R\overline{T}(x)}$ is the speed of sound.
\subsection{Calculating $M$:}
\begin{equation}
    M = \frac{\overline{u}(x)}{\overline{c}(x)},
\end{equation}
where $\overline{u}(x)$ is obtained by solving Eq.~(\ref{Eq:37})
\begin{equation}
    a_1\overline{u}^2(x)+a_2\overline{u}(x)+a_3\overline{T}(x) = 0. \label{Eq:B3}
\end{equation}
Here,
\begin{align}
    a_1 &= \overline{\rho}_0\overline{u}_0, \\
    a_2 &= -(\overline{p}_0+\overline{\rho}_0\overline{u}_0^2), \\
    a_3 &= \frac{\overline{p}_0\overline{u}_0}{\overline{T}_0}.
\end{align}

\subsection{Calculating $\alpha$:}
Consider the perfect gas law
\begin{equation}
    \overline{p} = \overline{\rho}R\overline{T}.
\end{equation}
Upon differentiating with respect to $x$, it can be written as
\begin{equation}
    \frac{1}{\overline{p}}\frac{d\overline{p}}{dx}-\frac{1}{\overline{\rho}}\frac{d\overline{\rho}}{dx} = \frac{1}{\overline{T}}\frac{d\overline{T}}{dx}.
\end{equation}
By substituting the expression for $d\overline{p}/dx$ from Eq.~(\ref{Eq:15}) and making use of the relations $M=\overline{u}/\overline{c}$ and $\overline{c}=\sqrt{\gamma R\overline{T}}$, the expression for $\alpha$ can be written as
\begin{equation}
    \alpha = \frac{1}{(\gamma M^2-1)}\frac{1}{\overline{T}}\frac{d\overline{T}}{dx}. \label{Eq:B9}
\end{equation}

\subsection{Calculating $dM/dx$:}
Consider the relation
\begin{equation}
    M = \frac{\overline{u}}{\overline{c}}.
\end{equation}
Differentiation with respect to $x$ gives the relation
\begin{equation}
    \frac{dM}{dx} = \frac{1}{\overline{c}}\frac{d\overline{M}}{dx}-\frac{M}{\overline{c}}\frac{d\overline{c}}{dx}.
\end{equation}
By substituting the expression for $d\overline{u}/dx$ from Eq.~(\ref{Eq:14}), and making use of Eq.~(\ref{Eq:B9}) and the relation $\overline{c}=\sqrt{\gamma R\overline{T}}$, the expression for $dM/dx$ can be written as
\begin{equation}
    \frac{dM}{dx} = -\frac{M\alpha}{2}(1+\gamma M^2).
\end{equation}

\subsection{Calculating $\beta$:}
From the perfect gas law ($\overline{p}=\overline{\rho}R\overline{T}$), it can be written that 
\begin{equation}
    \frac{d\overline{\rho}}{dx} = \frac{1}{R}\frac{1}{\overline{T}}\frac{d\overline{p}}{dx}-\frac{\overline{p}}{\overline{T}^2}\frac{d\overline{T}}{dx}.
\end{equation}
Upon successive differentiation with respect to $x$, the expression for $\beta$ can be obtained as
\begin{equation}
    \beta = \frac{1}{\overline{\rho}}\frac{d^2\overline{\rho}}{dx^2} = \frac{1}{\overline{p}}\frac{d^2\overline{p}}{dx^2}+2\left(\frac{1}{\overline{T}}\frac{d\overline{T}}{dx}-\frac{1}{\overline{p}}\frac{d\overline{p}}{dx}\right)\left(\frac{1}{\overline{T}}\frac{d\overline{T}}{dx}\right)-\frac{1}{\overline{T}}\frac{d^2\overline{T}}{dx^2}.
\end{equation}
Here, $d\overline{p}/dx$ can be evaluated from Eq.~(\ref{Eq:38}) as
\begin{equation}
    \frac{d\overline{p}}{dx} = a_1\frac{d\overline{u}}{dx} = -a_1\alpha\,\overline{u},
\end{equation}
and the expression for $d^2\overline{p}/dx^2$ can be evaluated from Eq.~(\ref{Eq:B3}) as
\begin{equation}
    \frac{d^2\overline{p}}{dx^2} = -\frac{2a_2\alpha^2\overline{u}^2+a_3\left(d^2\overline{T}/dx^2\right)}{2a_1\overline{u}+a_2}.
\end{equation}
\end{document}